\documentclass[preprint,12pt]{clear2024_for_arxiv} 

\title[Scalable Counterfactual Distribution Estimation in Multivariate Causal Models]{Scalable Counterfactual Distribution Estimation \\in Multivariate Causal Models}
\usepackage{times}
\usepackage{comment}

\usepackage{amsmath,amsfonts,bm}

\def\eqref#1{equation~\ref{#1}}

\def\1{\bm{1}}

\DeclareMathAlphabet{\mathsfit}{\encodingdefault}{\sfdefault}{m}{sl}
\SetMathAlphabet{\mathsfit}{bold}{\encodingdefault}{\sfdefault}{bx}{n}

\newcommand{\R}{\mathbb{R}}

\usepackage{url}
\usepackage{graphicx}
\usepackage{dcolumn}
\usepackage{bm}
\usepackage{enumitem}
\usepackage{authblk}
\usepackage{amsmath}
\usepackage{amsfonts}
\usepackage{nameref}
\usepackage{mathtools}
\usepackage{bm}

\usepackage{color,xcolor}

\usepackage{hyperref}

\newcommand{\norm}[1]{\left\lVert#1\right\rVert}

\makeatletter
\def\munderbar#1{\underline{\sbox\tw@{$#1$}\dp\tw@\z@\box\tw@}}
\makeatother

\newcommand{\be}{\begin{equation}}
\newcommand{\ee}{\end{equation}}
\newcommand{\bea}{\begin{equation*}\begin{aligned}}
\newcommand{\eea}{\end{aligned}\end{equation*}}

\newcommand{\mbb}{\mathbb}

\newcommand{\calO}{{\mathcal O}}

\newcommand{\RR}{\mbb R}

\clearauthor{\Name{Thong {Pham}} \Email{thong-pham@biwako.shiga-u.ac.jp}\\ \addr{\emph{Data Science and AI Innovation Research Promotion Center, Shiga University, Japan}}
\\
\addr \emph{Center for Advanced Intelligence Project, RIKEN, Japan}\\
\Name{Shohei Shimizu} \Email{shohei-shimizu@biwako.shiga-u.ac.jp}\\
\addr{\emph{Graduate School of Data Science, Shiga University, Japan}}\\
\addr{\emph{Center for Advanced Intelligence Project, RIKEN, Japan}}\\ 
\Name{Hideitsu Hino} \Email{hino@ism.ac.jp}\\
\addr{\emph{The Institute of Statistical Mathematics}}\\ 
\addr{\emph{Center for Advanced Intelligence Project, RIKEN, Japan}}\\
\Name{Tam Le} \Email{tam@ism.ac.jp}\\
\addr{\emph{The Institute of Statistical Mathematics}}\\
\addr{\emph{Center for Advanced Intelligence Project, RIKEN, Japan}}}

\begin{document}

\maketitle

\begin{abstract}%
  We consider the problem of estimating the counterfactual joint distribution of multiple quantities of interests (e.g., outcomes) in a multivariate causal model extended from the classical difference-in-difference design. Existing methods for this task either ignore the correlation structures among dimensions of the multivariate outcome by considering univariate causal models on each dimension separately and hence produce incorrect counterfactual distributions, or poorly scale even for moderate-size datasets when directly dealing with such multivariate causal model. We propose a method that alleviates both issues simultaneously by leveraging a robust latent one-dimensional subspace of the original high-dimension space and exploiting the efficient estimation from the univariate causal model on such space. Since the construction of the one-dimensional subspace uses information from all the dimensions, our method can capture the correlation structures and produce good estimates of the counterfactual distribution. We demonstrate the advantages of our approach over existing methods on both synthetic and real-world data.   
\end{abstract}


\section{Introduction}

Causal inference 
has received explosive interest in the last decades, due to the need to extract causal knowledge from  data in various research fields, such as statistics~\citep{causal_review_pearl_2009}, sociology~\citep{causal_sociology}, biomedical informatics~\citep{ causal_medical_IT}, public health~\citep{causal_public_health}, and machine learning~\citep{scholkopf_causality_review_2022}.
One of the most popular causal inference models in practice 
is the difference in difference (DiD) model, which dates back to the works of John Snow in the 1850s ~\citep{snow_1854, snow_1855,did_review_1,did_review_2011,did_review_3}. 
In this model, we observe a quantity of interest (i.e., outcome) from two different groups, i.e., the treatment group and the control group, at two different time steps, i.e., before and after the intervention (i.e., treatment) event. More precisely, the intervention is only applied on the treatment group while there is no intervention applied on the control group. We assume that if there was no intervention in the treatment group, its outcome variable would evolve the same way as that of the control group. This is the so-called ``\emph{parallel trend}'' assumption. The classical DiD model, however, requires that the parallel trend is linear, i.e., the means of the outcome variable in the two groups must evolve the same way when there is no intervention. This may limit its application in practice.
To extend the DiD model for non-linear settings, several proposals have been developed in the literature~\citep{Abadie_2006,athey_imbens_2006,blundell_costa_2009,Sofer_2016}, of which notably is the Changes-in-Changes (CiC) model~\citep{athey_imbens_2006}.

The CiC model generalizes the DiD model to include non-linear parallel trends that can act on the whole distribution of the outcome, i.e., in the absence of intervention, the means of the outcome in control and treatment groups are allowed to evolve in different ways, as long as the two outcome \emph{distributions} evolve in the same way. This allows identifications of more complex treatment effects that require information from the whole outcome distributions, not just their first moments~\citep{did_review_2011}. The standard CiC model, however, is only designed for univariate outcomes. In order to extend the CiC model for a multivariate outcome variable, a naive approach is to tensorize univariate CiC models, i.e., considering independently a univariate CiC model for each coordinate. Nevertheless, this naive approach fails to capture correlations among coordinates of the outcome and thus is incapable of modelling complex, multivariate parallel trends.

Recently, by leveraging the optimal transport (OT) theory~\citep{Villani-03, Villani-09},~\citet{torous_2021} proved that the counterfactual outcome distribution of the treatment group in the CiC model (i.e., the outcome distribution of the treatment group \emph{without} receiving intervention) at the post-intervention time stamp can be estimated by exploiting the \emph{optimal transport map} which pushforwards the outcome distribution of the control group at the pre-intervention time stamp to that at the post-intervention time stamp. Consequently, it is natural to extend the CiC model for univariate quantity of interest into that for multivariate one through the lens of OT, since this would take into account the dependence structure of the dimensions and be able to model complex parallel trends.

However, OT suffers a few drawbacks. It has a high computational complexity, which is super cubic with respect to the number of supports of the input distribution. A popular approach is to rely on the entropic regularization for OT, a.k.a., Sinkhorn~\citep{Cuturi-2013-Sinkhorn}, to reduce its computational complexity to quadratic. Yet, Sinkhorn yields a dense estimator for the optimal transport plan, which is not a desirable property for counterfactual estimation in the CiC model. Additionally, OT has a high sample complexity, i.e., $\calO(n^{-1/d})$ where $n$ is the number of samples and $d$ is the dimension of samples in the probability measures.

In this work, in order to exploit the efficient computation of the CiC model for the univariate quantity of interest, and alleviate the above-mentioned challenges of the OT approach for the multivariate CiC model, we propose to leverage the max-min robust OT approach~\citep{pmlr-v97-paty19a}. In particular, we propose to lift the univariate CiC model to that for a multivariate quantity of interest by seeking a \emph{robust} latent univariate subspace. Unlike the naive tensorization approach, our approach can incorporate the correlations of coordinates. Moreover, unlike the standard OT approach as in~\citep{torous_2021}, our estimator can preserve the efficient computation as in the univariate CiC model since the optimal transport plan is estimated on the robust latent one-dimensional subspace instead of its original high-dimensional space. 

Intuitively, our approach follows the \emph{max-min robust OT} approach~\citep{pmlr-v97-paty19a},  which steams from the robust optimization~\citep{ben2009robust, bertsimas2011theory} where there are uncertainty non-stochastic parameters. The robust optimization has many roots in applied sciences, e.g., in robust control~\citep{keel1988robust}, machine learning~\citep{morimoto2001robust, xu2009robustness, NEURIPS2022_robustRL}. In the context of OT, several advantages of the max-min (and its relaxation min-max) robust OT have been reported. For example, (i) it makes the OT approach robust to noise~\citep{pmlr-v97-paty19a, dhouib2020swiss}; and (ii) it also helps to reduce the sample complexity~\citep{pmlr-v97-paty19a, Deshpande_2019_CVPR}. At a high level, our contributions are two-fold as follows:
\begin{itemize}
    \item (i) We propose a max-min robust OT approach for the multivariate CiC model. The proposed approach not only inherits properties of the OT approach for the CiC model but also preserves the efficient computation as in the univariate CiC model.
    \item (ii) We evaluate our approach on both synthesized and real data to illustrate the advantages of the proposed method.
\end{itemize}

The paper is organized as follows. After reviewing the multivariate CiC model and existing methods for estimating the counterfactual distribution in this model in Section~\ref{sec:causal_model}, we discuss our proposed method in Section~\ref{sec:proposed_method} and demonstrate its benefits through synthetic data in Section~\ref{sec: simulation}. We apply it to the classical dataset of~\cite{NBERw4509} in Section~\ref{sec: real_data} before giving concluding remarks in Section~\ref{sec: concluding}.

\textbf{Notations.} We use the superscripts \texttt{C} and \texttt{T} to indicate the control group and treatment group, respectively. We drop those superscripts when either the context is clear or it is not necessary to distinguish these two groups.


\section{The Causal Model}\label{sec:causal_model}
In this section, we describe the CiC causal model for multiple quantities of interests~\citep{torous_2021}, which is an extension based on OT theory from the original, univariate model~\citep{athey_imbens_2006}.

We use a stochastic process to model the quantity of interests (i.e., outcomes) before the intervention, i.e., at the time stamp $t=0$, and after the intervention, i.e., at the time stamp $t=1$. We denote them as $\{Y_t\}_{t=0, 1}$ where $Y_t$ is in the $\RR^d$ space. For the original CiC causal model, we have $d=1$~\citep{athey_imbens_2006}. We let $\mu_t$ be the distribution of $Y_t$ for $t = 0, 1$. Without loss of generality, we assume that $Y_t$ is generated from a latent variable $U_t \in \RR^d$ which may change over time, but the distribution $\nu$ of the latent variable $U_t$ is not changed over time, i.e., time-invariant $U_t \sim \nu$ for $t=0, 1$. Intuitively, $U_t$ may be regarded as (unobserved) intrinsic features of a sample.

In the CiC model, we observe two groups:
\begin{itemize}
    \item (i) the control group: the stochastic process $\{Y_t^{\texttt{C}}\}$ is solely affected by the \emph{natural drift}. The evolution from $\{Y_0^{\texttt{C}}\}$ at $t = 0$ to $\{Y_1^{\texttt{C}}\}$ at $t = 1$ is independent of the treatment effects (of intervention).
    \item (ii) the treatment group: the stochastic process $\{Y_t^{\texttt{T}}\}$ is affected by both the \emph{natural drift} and the \emph{treatment effects}.
\end{itemize}

For the CiC causal model, the goal of our causal inference is to deconvolve the \emph{natural drift} and the \emph{treatment effects} in the treatment group. For example, we would like to estimate the counterfactual distribution of the control group at post-intervention under \emph{only natural drift effect} (i.e., without the treatment effects). By doing so, we can estimate the \emph{treatment effects} of the intervention for the considered groups in application domains.
    
\subsection{The Natural Drift Model}\label{subsec:natural_drift}

Natural drift is best explained in the stochastic process $\{Y_t^{\texttt{C}}\}$ for $t=\{0, 1\}$ since this process involves solely the natural drift and is not affected by the treatment effect. The change of $\{Y_t^{\texttt{C}}\}$ from the pre-intervention ($t=0$) to the post-intervention ($t=1$) is modeled by assuming the existence of two \emph{production functions} $h_t: \RR^d \mapsto \RR^d$ with $t \in \{0, 1 \}$ such that 
\[
Y_t^\texttt{C} = h_t\left(U_t^\texttt{C}\right).\] Consequently, we have $\mu_t^\texttt{C} = (h_t)_{\sharp}\nu^\texttt{C}$ where we introduce a new notation $\sharp$ as the \emph{pushforward} operator which is defined as for any measurable set $A \subseteq \RR^d$, $\nu(A) = \mu(h_t^{-1}(A))$~\citep[Def. 2.1]{peyre2019computational}. In other words, the distribution of the quantity of interests in the control group at the time stamp $t$ (i.e., $\mu_t^C$) is the pushforward of the distribution of the latent variable $U_t^{\texttt{C}}$ of the control group at the time stamp $t$ (i.e., $\nu^\texttt{C}$) by the production function $h_t$. 

\paragraph{Natural drift map.} Assume that the production function $h_0$ is invertible, we have $\nu^\texttt{C} = (h_0^{-1})_{\sharp}\mu_0^C$. Additionally, the distribution $\nu^\texttt{C}$ of $U_t^\texttt{C}$ is time-invariant. Thus, we have
\begin{equation}
    \mu_1^\texttt{C} = (h_1 \circ h_0^{-1})_{\sharp} \mu_0^\texttt{C}. \label{eq:pushforwar_control}
\end{equation}
Equivalently, $\mu_1^\texttt{C}$ is the pushforward of $\mu_0^\texttt{C}$ by the \emph{natural drift map} 
\begin{equation}\label{eq:natural_drift}
\texttt{f} = h_1 \circ h_0^{-1}.
\end{equation}

\paragraph{Natural drift in the treatment group.} We first introduce the concept of a \emph{counterfactual distribution} of the outcome in this group, which is the outcome variable of the treatment group at post-intervention $t = 1$ under the purely hypothetical situation that the treatment was never applied. Denote $\{Y_1^{\texttt{T}^*}\}$ as this hypothetical stochastic process for the outcome and its distribution as $\mu_1^{\texttt{T}^*}$. It is assumed that the change from $\{Y_0^\texttt{T}\}$ to $\{Y_1^{\texttt{T}^*}\}$ is governed by the same production functions $h_0$ and $h_1$ used in modeling natural drift in the control group, namely $Y_0^\texttt{T} = h_0(U_0^T)$ and $Y_1^{\texttt{T}^*} = h_1(U_1^T)$. This assumption generalizes the parallel trend assumption in the classical DiD model. Then, similarly to Eq.~(\ref{eq:pushforwar_control}), we have 
\begin{equation}
\mu_1^{\texttt{T}^*} = (h_1 \circ h_0^{-1})_{\sharp}\mu_0^\texttt{T} = (\texttt{f})_{\sharp} \mu_0^\texttt{T},\label{eq:pushforward_treatment}
\end{equation}i.e., the counterfactual distribution $\mu_1^
{\texttt{T}^*}$ is the pushforward of $\mu_{0}^\texttt{T}$ by the natural drift map $\texttt{f}$.

\begin{figure}[!h]
     \centering
\includegraphics{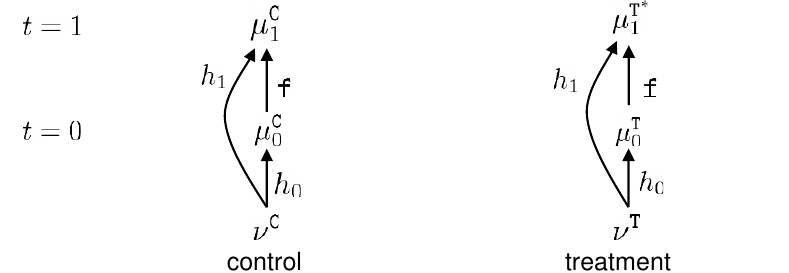}
\caption{\textbf{Schematic of the multivariate CiC model}. The latent distributions $\nu^\texttt{C}$ and $\nu^\texttt{T}$ are time-invariant. At each time $t$, the production function $h_t$ is applied to each latent distribution to produce the corresponding outcome distribution in the absence of intervention. Consequently, the natural drift map $\mathtt{f} = h_1\circ h_0^{-1}$ dictates the evolutions of the outcome distribution in both control and treatment groups. This generalizes the linear parallel trend assumption in the classical DiD model. For estimating treatment effects on the treatment group, our goal is to estimate $\mu_{1}^{\texttt{T}^{*}}$, the counterfactual distribution of outcomes in the treatment group at $t = 1$. One can estimate $\mathtt{f}$ from the observable empirical versions of $\mu_0^{\texttt{C}}$ and $\mu_1^{\texttt{C}}$ then use this estimation of $\mathtt{f}$ to push-forward the observable empirical version of $\mu_0^{\texttt{T}}$ to obtain an estimation of $\mu_{1}^{\texttt{T}^{*}}$.} 
\vspace{-2em}
\label{fig:causal_model}
\end{figure}

Equations~(\ref{eq:pushforwar_control}) and~(\ref{eq:pushforward_treatment}) suggest the following two-step method to estimate $\mu_1^{\texttt{T}^*}$, the counterfactual distribution of outcomes in the treatment group under only effects of the natural drift:
\begin{itemize}
    \item (i) We first estimate the natural drift map $\texttt{f}$ from observed samples in the stochastic processes $\{Y_{0}^\texttt{C}\}, \{Y_{1}^\texttt{C}\}$ of the distributions  $\mu_{0}^\texttt{C}, \mu_{1}^\texttt{C}$ respectively. How to perform this estimation will be discussed in the next section.
    
    \item (ii) We then use the estimated natural drift map $\texttt{f}$ as the pushforward function for the stochastic process $\{Y_{0}^\texttt{T}\}$ with distribution $\mu_{0}^{\texttt{T}}$ to obtain an estimate for the distribution $\mu_1^{\texttt{T}^*}$.
\end{itemize}
The schematic summary of the causal model is shown in Fig.~\ref{fig:causal_model}.

If we further assume $U_0^\texttt{C} = U_1^{\texttt{C}}$, then we have $Y_1^\texttt{C} = \texttt{f}\left(Y_0^\texttt{C}\right)$, i.e., the natural drift map can be estimated by regress the control group at $t = 1$ on the control group at $t = 0$. However, it requires coupled observations $\left(Y_0^{\texttt{C}}, Y_1^{\texttt{C}}\right)$, which may be not available in practical applications, e.g., single-cell RNA-Seq data.
\begin{remark}
For $x\in \RR^d$, denote $x^i$ ($i = 1,\cdots, d$) as the $i$-th coordinate of $x$. The multivariate CiC model is equivalent to the naive tensorization of univariate CiC models when (i) each production function $h_t: \RR^d \mapsto \RR^d$ ($t \in \{0,1\}$) can be decomposed as $h_t(x) = [h_t^1(x^1)\cdots h_t^d(x^d)]^T$ for univariate functions $h_t^i:\RR \mapsto \R$, and (ii) each coordinate of the latent variable (in both control and treatment groups) is independent. Therefore, it is difficult for the tensorization of univariate CiC models to express complex, multivariate natural drifts. 
\end{remark}

\paragraph{The CiC causal model and optimal transport.} For the uncoupled observations $Y_0^{\texttt{C}}, Y_1^{\texttt{C}}$, given $d = 1$, the original CiC estimator~\citep{athey_imbens_2006} assumes that $h_0$ and $h_1$ are \emph{monotone increasing}. Let $F_t^{\texttt{C}}$ be the cumulative distribution function (cdf) of $Y_t^{\texttt{C}}$ for $t = 0, 1$. Then, the unique monotone increasing \emph{natural drift map} is given by $\left(F_1^{\texttt{C}}\right)^{-1} \circ F_0^{\texttt{C}}$ such that $\mu_1^\texttt{C} = \left(\left(F_1^{\texttt{C}}\right)^{-1} \circ F_0^{\texttt{C}}\right)_{\sharp} \mu_0^{\texttt{C}}$. Interestingly, the natural drift map $\left(\left(F_1^{\texttt{C}}\right)^{-1} \circ F_0^{\texttt{C}}\right)$ is also the optimal transport map between $\mu_0^{\texttt{C}}$ and $\mu_1^{\texttt{C}}$. Therefore, the OT theory provides a natural framework to extend the Changes-in-Changes causal model for multivariate outcomes~\citep{torous_2021}.

\subsection{The OT approach for estimating the counterfactual distribution in multivariate CiC}\label{sec:sub_from_uni_to_high}


As briefly discussed in Section~\ref{subsec:natural_drift}, for the univariate outcome, the OT map is the natural drift map estimation for CiC causal model.~\citet{torous_2021} leveraged the OT theory to extend it for the multivariate quantity of interests. We will discuss some mathematical background of OT for estimating $\texttt{f}$ in high dimensional setting for the CiC causal model. 

Let $\mu, \nu$ be two probability measures supported on $\RR^d$, the OT problem between $\mu$ an $\nu$ with squared Euclidean ground cost is defined as
\begin{equation}\label{eq:OT}
\text{OT}(\mu, \nu) = \min_T \int_{\RR^d} \norm{x - T(x)}_2^2 \, \text{d}\mu \quad \text{s.t.} \quad (T)_{\sharp}\mu = \nu,
\end{equation}
where $T$ is known as the transport plan.

\paragraph{For univariate model ($d=1$).} Given a probability measure $\mu$ supported on $\RR$, we define its cumulative distribution function (cdf) $F_{\mu}$ as
\begin{equation}
F_{\mu}(x) = \mu((-\infty, x]).
\end{equation}
Note that the cdf is not always invertible, since it is not strictly increasing. The pseudo-inverse of cdf $F^{-1}: [0, 1] \mapsto \RR$ is given by
\begin{equation}
    F^{-1}(x) = \inf \left\{ t \in \RR \mid F(t) \ge x \right\}. 
\end{equation}
When $d=1$, the OT admits a closed-form expression for the optimal map $T$ as follows:
\begin{equation}
    T = \left(F_{\nu}\right)^{-1} \circ F_{\mu}.
\end{equation}
Note that $T$ is the unique increasing map such that $(T)_{\sharp}{\mu} = \nu$.\footnote{The optimal condition for OT (i.e., existence of the optimal map $T$) was described in \citep{gangbo1999monge} for $d=1$.} Therefore, if $\mu = \mu_{0}^{\texttt{C}}$, $\nu = \mu_{1}^{\texttt{C}}$, and $\texttt{f}$ is increasing, e.g., by choosing monotone $h_0$ and $h_1$, then $\texttt{f}$ is the OT map between $\mu_{0}^{\texttt{C}}$
 and $\mu_{1}^{\texttt{C}}$.

\paragraph{For multivariate model ($d>1$).} The natural drift map $\texttt{f}$ can be estimated via the OT map between $\mu_0^{\texttt{C}}$ and $\mu_1^{\texttt{C}}$~\citep{torous_2021}.\footnote{The optimal condition for OT (with $d>1$) was derived in \citep{brenier1991polar}.} However, for high-dimensional space, OT suffers a few drawbacks: (i) computational complexity, i.e., super cubic $\calO(n^3\log(n))$ where $n$ is the number of supports of input measures; (ii) high sample complexity, i.e., $\calO(n^{-1/d})$, which requires too many samples to precisely estimate the OT between two continuous distributions (e.g., $\mu_0^{\texttt{C}}, \mu_1^{\texttt{C}}$).

A naive approach to estimate the natural drift map in the multivariate CiC causal model is to decompose the model for each dimension. Specifically, one treats the multivariate CiC causal model with $d$-dimensional outcomes ($d>1$) as $d$ independent univariate CiC causal models, a.k.a., the tensorization of univariate CiC causal models. It is then efficient to estimate univariate OT maps for these univariate CiC causal models via their closed-form expressions of the corresponding univariate OT problems as in the original CiC causal model~\citep{athey_imbens_2006} instead of solving the high-complexity full OT problem for measures supported in high-dimensional spaces~\citep{torous_2021}. One then tensorizes all the estimated univariate OT maps to create an estimation of the multivariate OT map. However, this approach might fail to capture the dependence structure among dimensions of the multivariate OT map, i.e., the natural drift map $\texttt{f}$, and thus when one uses this tensorized map to pushforward samples of $\mu_0^{\texttt{T}}$, one might produce a counterfactual distribution with a wrong dependence structure (e.g., see Fig.~\ref{fig:illustrative_example}).

In this work, we propose an efficient approach to leverage the advantages of the univariate CiC causal model for the multivariate CiC by seeking a robust latent $1$-dimensional space for OT estimation. More precisely, our approach is inspired by the subspace robust OT~\citep{pmlr-v97-paty19a} whose authors proposed to estimate the OT map in low-dimensional subspace to reduce the sample complexity for the OT problem, and further increase the robustness of OT estimation with respect to noise.

\section{The proposed method based on robust OT over latent $1$-dimensional subspaces}\label{sec:proposed_method}

In order to efficiently leverage the advantages of the univariate CiC causal model and mitigate issues in the naive tensorization for the multivariate CiC, in this work, we propose to seek a robust latent $1$-dimensional subspace as a surrogate to estimate the univariate OT map to bridge the univariate and multivariate CiC causal models. This approach is also known as \emph{max-min robust variant of OT}~\citep{pmlr-v97-paty19a}.

Let $\mathcal{G}_1$ be the Grassmannian of the $1$-dimensional subspaces of $\RR^d$, defined as
\[
\mathcal{G}_1 = \{E \subset \RR^d \mid \text{dim}(E) = 1 \},
\]
where $\text{dim}(E)$ is the dimension of the space $E$. Given two measures $\mu, \nu$ supported on the space $\RR^d$, the max-min robust OT~\citep{pmlr-v97-paty19a} considers the maximal OT distance over all possible $1$-dimensional projections of input probability measures. Denote $P_{E}$ as the projector on the $1$-dimensional space $E$, the robust OT is defined as
\begin{equation}\label{eq:robustOT}
\widetilde{\text{ROT}}(\mu, \nu) = \sup_{E \in \mathcal{G}_1} \text{OT}({P_{E}}_{\sharp}\mu, {P_{E}}_{\sharp}\nu).
\end{equation}
One way to parameterize the projection on the $1$-dimensional space $P_E$ and the Grassmannian manifold $\mathcal{G}_1$ is to utilize a projected direction vector $\omega$ on the sphere centering at the origin with a radius $1$ as follows:
\begin{equation}\label{eq:max1dOT}
\overline{\text{ROT}}(\mu, \nu) = \max_{\omega \in \RR^d \, | \, \norm{\omega}_2 = 1} \text{OT}({P_{\omega}}_{\sharp}\mu, {P_{\omega}}_{\sharp}\nu),
\end{equation}
where $P_{\omega}$ denotes the projector on the direction $\omega$. For a support $x \in \RR^d$, its projection by the projected direction $\omega$ is computed by $\left< x, \omega\right>$. However, the problem $\overline{\text{ROT}}$ is non-convex, which is usually approximated by the first-order method in practical applications~\citep{Deshpande_2019_CVPR}.\footnote{$\overline{\text{ROT}}$ is also known as the max-sliced Wasserstein~\citep{Deshpande_2019_CVPR}.}

Much as the sliced-Wasserstein (SW)~\citep{rabin2011wasserstein}\footnote{SW projects supports into $1$-dimensional space and exploits the closed-form expression of the univariate OT.} which is usually approximated by averaging over a few random directions in practical applications instead of integrating over all possible directions on the sphere, and in order to optimize the robust OT efficiently, we also seek a direction over a subset of projected directions $\Omega$ for the robust OT, defined as follows:
\begin{equation}\label{eq:ROT}
\text{ROT}(\mu, \nu) = \max_{\omega \in \Omega} \, \text{OT}({P_{\omega}}_{\sharp}\mu, {P_{\omega}}_{\sharp}\nu),
\end{equation}
where we construct $\Omega$ by randomly sampling $k$ directions similar as SW. We observe that a small $k$, e.g., in the range of $10$ to $50$, provides a good balance between computation speed and accuracy. For estimating the natural drift map $\texttt{f}$, $\mu$ and $\nu$ are chosen to be the empirical versions of $\mu_{0}^{\texttt{C}}$ and $\mu_{1}^{\texttt{C}}$ respectively.

While our approach inherits the fast computation of the CiC estimator for univariate causal models and thus can scale up for large-scale causal inference applications, it also gives good performances in estimating the natural drift map $\texttt{f}$. This is in line with previous observations from various applications of sliced-based OT~\citep{pmlr-v97-paty19a, Deshpande_2019_CVPR}.

\begin{remark}
A technical remark is that it may not exist the optimal transport map $T$ for the OT problem in Eq.~(\ref{eq:OT}) (i.e., the Monge formulation of OT problem). In practical applications, one can relax such the Monge problem of OT by the Kantorovich formulation of OT problem~\citep{peyre2019computational}, in which the optimal transport plan always exists.\footnote{We give a review about the Kantorovich formulation of OT problem in Section~\ref{sec:supp_review_OT}.} Consequently, one can utilize such optimal transport plan and leverage the barycentric map~\citep{bonneel2016wasserstein} for the OT map in the Monge problem of OT.
\end{remark}

\section{Synthetic experiments}
\label{sec: simulation}
All experiments were carried out on commodity hardware and can be reproduced by the code in the supplementary. From here, we will denote the naive tensorization of the univariate CiC method for estimating multidimensional counterfactual distributions simply as the CiC method.

\subsection{Illustrative examples}\label{sec:sub_illustrative}
We demonstrate our method through two 2D examples for ease of visualization. For the latent distributions $\nu^{\texttt{C}}$ and $\nu^{\texttt{T}}$, we choose bivariate Gamma distributions in the first example and 2D Gaussian mixtures in the second example. These settings lead to an unimodal counterfactual for the Gamma case, and a multimodal counterfactual distribution for the Gaussian mixture case, as can be seen in the ground truth panels of Fig.~\ref{fig:illustrative_example}. More details on experiment settings can be found in Appendix~\ref{appendix:sim_settings}.  

\begin{figure}[!h]
\centering
\includegraphics[width = \textwidth]{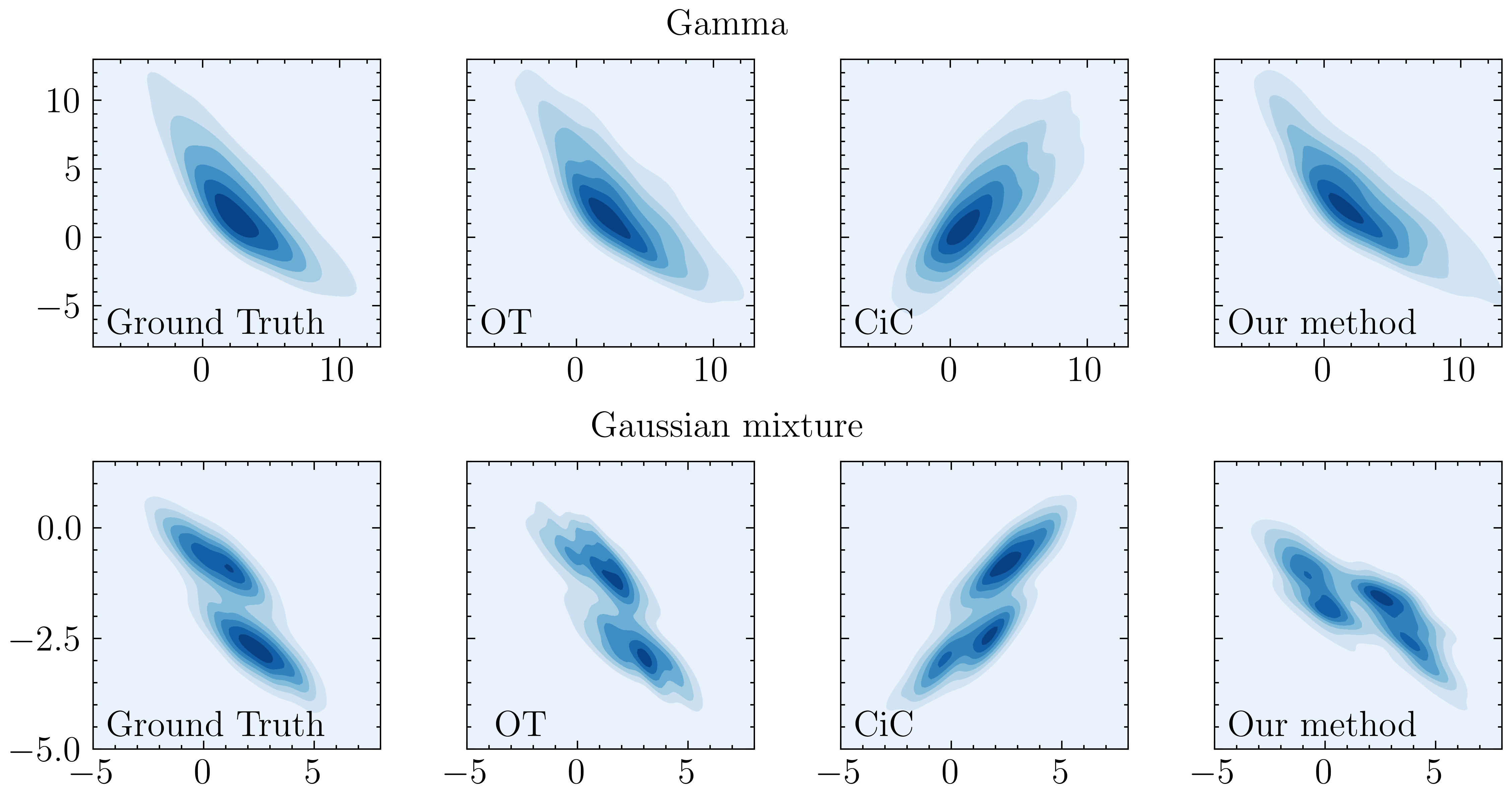}
\caption{\textbf{Counterfactual distribution estimated by each method}. While CiC failed to capture the dependence structures between dimensions of the counterfactual distributions in both examples, our method succeeded.}
\label{fig:illustrative_example}
\vspace{-1em}
\end{figure}

The proposed method produced a counterfactual distribution with the correct correlation structure, while CiC failed to do so. This demonstrates the 1-dimensional subspace created by our method can indeed provide the type of information that CiC cannot capture. To systematically evaluate the performance of each method, we generate $10$ datasets and then measure the running time as well as the OT distance between the estimated counterfactual distribution of each method and the empirical version of the true counterfactual distribution $\mu_{1}^{\texttt{T}^*}$. This empirical distribution, denoted as $\widetilde{\mu_{1}^{\texttt{T}^*}}$, is regarded as the ground truth in our experiments. The averaged values are reported in Fig.~\ref{fig:illustrative_example}.   

\begin{figure}[!h]
\centering
\includegraphics[width = \textwidth]{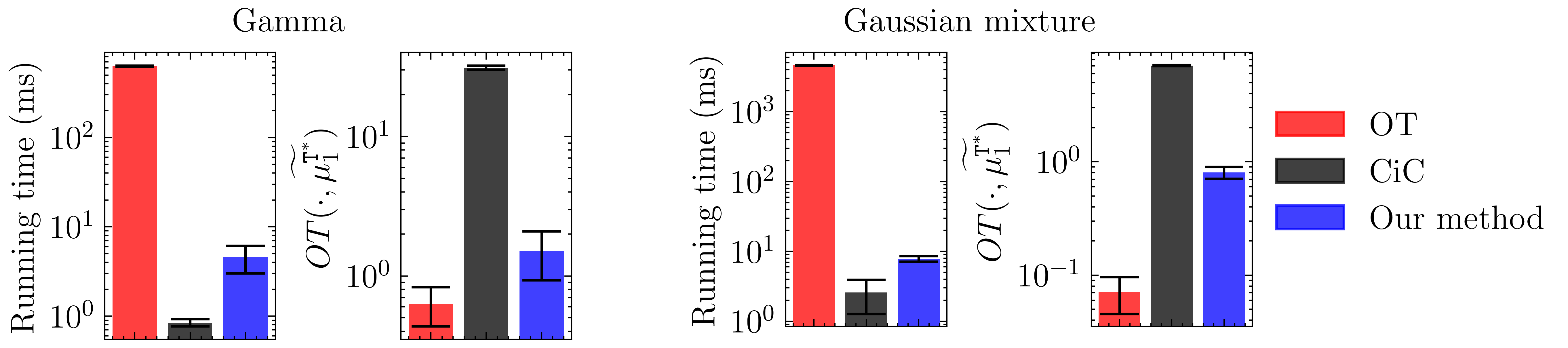}
\caption{\textbf{Averaged running time and OT distance to ground truth}. The OT distance is measured between the estimation result of each method and the ground truth $\widetilde{\mu_{1}^{\texttt{T}^*}}$, which is the empirical version of the true counterfactual distribution. Values are calculated over ten datasets. Our method outperforms the OT approach in terms of speed while being significantly better than the CiC approach in terms of accuracy measured by OT distance to ground truth.} 
\vspace{-2em}
\label{fig:illustrative_example_bar_plot}
\end{figure}

The CiC estimator has the worst averaged OT distance to ground truth among the three methods. This might be due to its failure to capture the correlation structure between dimensions. The proposed method is about one order smaller in OT distance to ground truth than the CiC while being about two orders faster than the OT approach. 

\subsection{Varying the number of samples $n$}
\begin{figure}[!h]\centering
     \centering
\includegraphics[width = \textwidth]{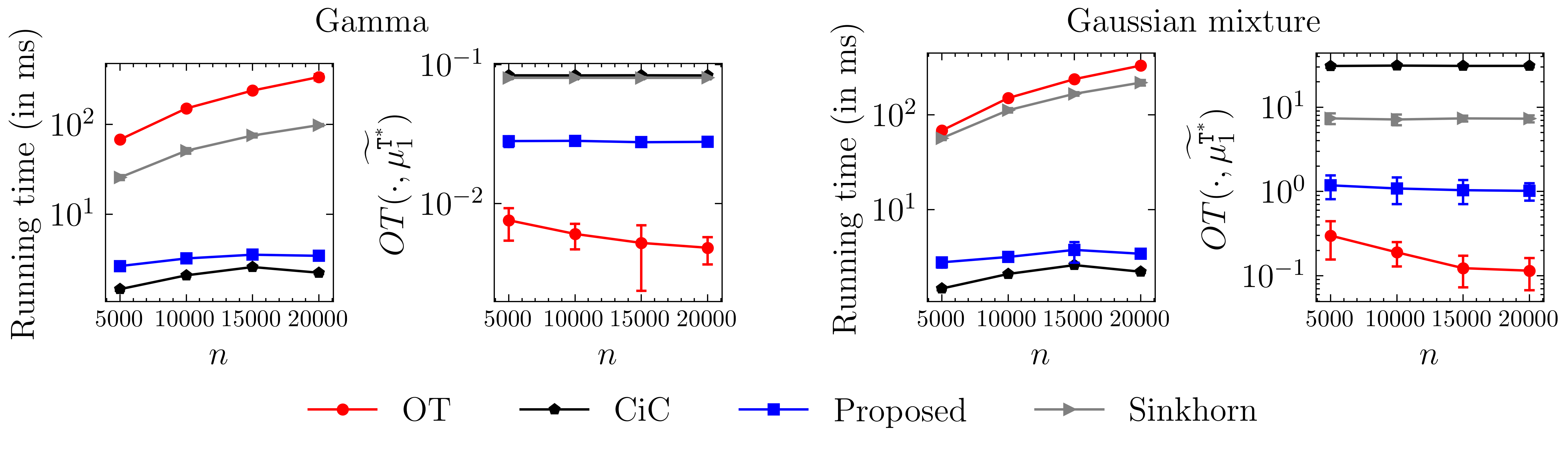}
         \caption{\textbf{Effect of varying the number of samples $n$ for two types of latent distributions}. In both cases, the proposed method is much faster than both the OT and Sinkhorn approaches, while being much more accurate than the CiC.}
         \vspace{-2em}
        \label{fig:experiment_varying_n}
\end{figure}

In this experiment, we check the findings in the illustrative examples by running the same experiment setting, with $d$ fixed at $2$, for various values of $n$. For each value of $n$, we generate $10$ datasets and report the averaged running time as well as the averaged OT distance between the estimated counterfactual distribution of each method and the ground truth $\widetilde{\mu_{1}^{\texttt{T}^*}}$. We also add Sinkhorn as a baseline. The results are shown in Fig.~\ref{fig:experiment_varying_n}.  

The running time of our method is close to that of the CiC and is faster than OT and Sinkhorn for all values of $n$. This is in line with the theoretical worst-case running time of each method. Regarding OT distance to ground truth, while being worse than OT, our method outperforms CiC for all values of $n$. This suggests that, while a large $n$ might help CiC in estimating the marginals of the counterfactual distribution in each dimension, the advantages of the 1-dimensional subspace constructed by our method do not diminish. 

In comparison with Sinkhorn, our method is both faster and more accurate. We caution that the performance of Sinkhorn depends heavily on the strength of the entropic regularization term, and choosing a suitable value for the entropic regularization hyperparameter is non-trivial. However, as stated earlier, since the worst-case running time of Sinkhorn is quadratic, it is reasonable to expect it to be generally slower than our method. Additional results that include Sinkhorn with different hyperparameters are shown in Appendix~\ref{appendix:sinkhorn}.

\subsection{Varying the dimension $d$} 
\begin{figure}[!h]
     \centering
         \includegraphics[width=0.7\textwidth]{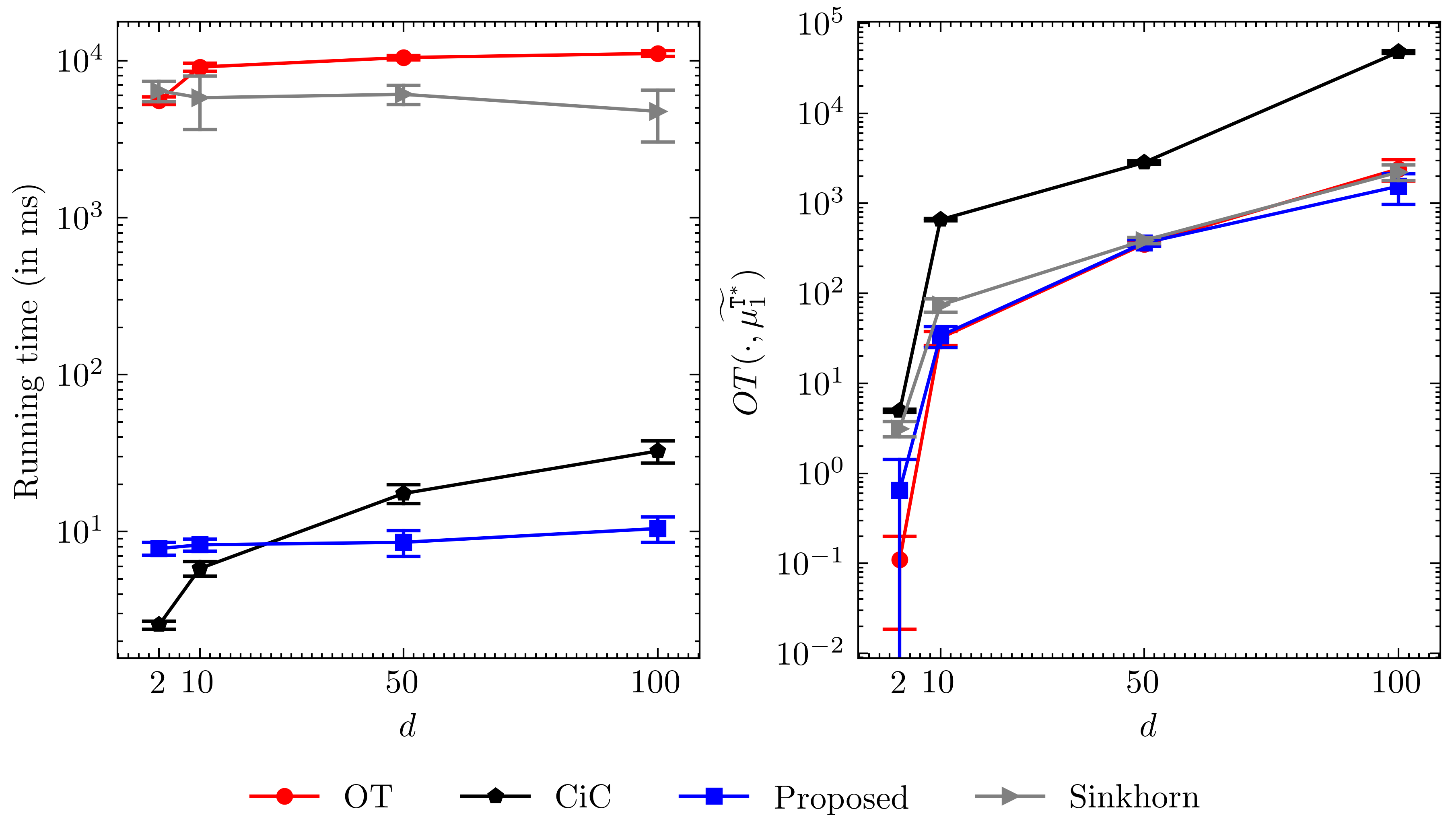}
         \caption{\textbf{Varying the dimension $d$ while fixing number of samples $n = 5000$}. The latent distribution is multivariate Gamma. When $d$ is high, our method is among the bests in both running time and accuracy. This suggests that the robust 1-dimensional subspace of our method can capture meaningful information even in high-dimensional situations.}
         \vspace{-2em}
        \label{fig:experiment_varying_d}
\end{figure}

In order to preserve the computational efficiency of CiC, the robust subspace in our method has to be one-dimensional. We investigate whether this subspace can still capture meaningful information in high-dimensional cases, by varying $d$ while keeping the number of samples $n$ fixed at $5000$. In this experiment, the latent distribution is multivariate Gamma. For each $d$, we generate $10$ datasets and report the averaged running time as well as the averaged OT distance to ground truth. More details on experiment settings can be found in Appendix~\ref{appendix:sim_settings}. The results are shown in Fig.~\ref{fig:experiment_varying_d}. Additional results that include Sinkhorn with different hyperparameters are shown in Appendix~\ref{appendix:sinkhorn}.

It is interesting to observe that when $d$ is high, our method is the fastest method, i.e., even faster than CiC. This is in line with the theoretical worst-case running time of CiC being $\mathcal{O}(dn \log n)$ and ours being $\mathcal{O}(n \log n + dn)$ when $k$ is fixed. As $d$ increases, the superiority of OT over our method in terms of OT distance to the ground truth diminishes and even reverses: our method is as accurate as, if not better than, OT when $d = 100$. Since the performance of CiC is still much worse, this reversal comes from the robust one-dimensional subspace. This is consistent with previous observations that found robustifications can alleviate the poor sample complexity of OT~\citep{pmlr-v97-paty19a}.

\subsection{Varying the number of projections $k$}
\begin{figure}[!h]
     \centering
         \includegraphics[width=\textwidth]{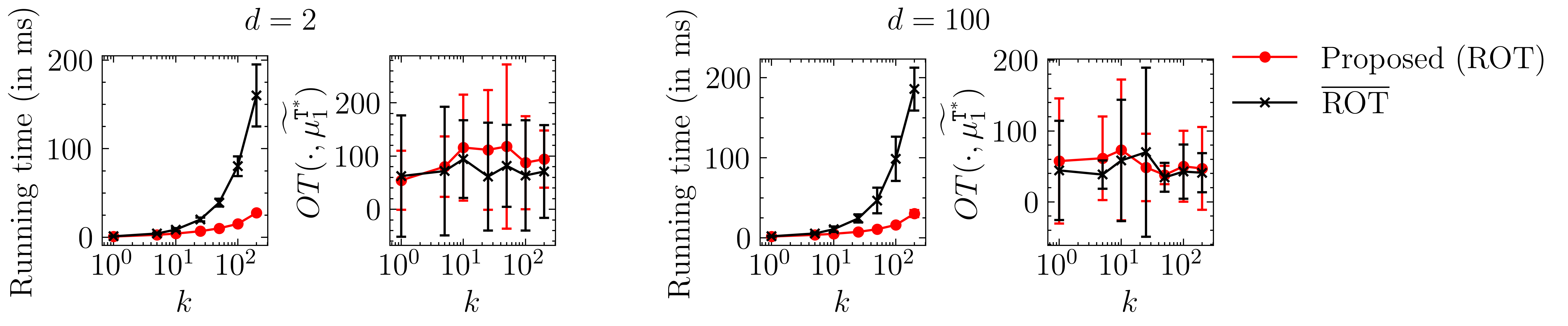}
         \caption{\textbf{Performances of our proposed method and an approach employed the $\overline{\text{ROT}}$ function when varying $k$}. Here $k$ has two different meanings depending on which method is considered. In our proposed method that uses the $\text{ROT}$ function in Eq.~(\ref{eq:ROT}), $k$ is the number of projections used to construct the robust one-dimensional subspace. In the approach that employs the $\overline{\text{ROT}}$ function in Eq.~(\ref{eq:max1dOT}), $k$ is the number of iterations of Adam first-order method. Our approach is as good as $\overline{\text{ROT}}$ in terms of accuracy while being much faster. Increasing the number of projections in our method to the hundreds marginally improves the accuracy and variance of the results, while requiring more time.}
         \vspace{-1.5em}
\label{fig:experiment_varying_num_of_projection}
\end{figure}

In all previous experiments, the number of projections used in constructing the robust 1-dimensional subspace has been fixed at $k = 10$. Using the same setting as the experiment above, for each $d$, we look at one dataset and inspect how varying $k$ affects the quality of the method. One trade-off to expect is that increasing $k$ might improve the accuracy of the estimation, e.g., finding a better subspace or reducing the variance between each run, at the cost of longer running time. We also investigate a closely related approach to estimate the counterfactual distribution by using the $\overline{\text{ROT}}$ objective function  in Eq.~(\ref{eq:max1dOT}) optimized by Adam~\citep{KingBa15} with different numbers of iterations. The results are shown in Fig.~\ref{fig:experiment_varying_num_of_projection}.

We found that increasing $k$ to the hundreds indeed improves the variance and accuracy of the result of our method, as can be seen from the mean and variance of OT distance to ground truth when $d = 100$. However, these improvements are marginal and come at the cost of a linearly longer running time. This is the reason we suggest choosing a small value for $k$, namely around $10-50$, as a reasonable balance region between accuracy and running time. We also observe that the $\overline{\text{ROT}}$ approach with Adam offered no improvement in terms of variance and accuracy of the results, while significantly running longer. This ineffectiveness of Adam might be due to the non-convexity and non-smooth of $\overline{\text{ROT}}$. 

\section{A real dataset example}
\label{sec: real_data}
We demonstrate the working of our proposed method on the classical data of~\citet{NBERw4509} (CK). On April 1, 1992, New Jersey's minimum wage rose from \$4.25 to \$5.05 per hour, while Pennsylvania's did not. This provided an opportunity to estimate the causal impact of the rise on employment in fast-food restaurants in
New Jersey, by analyzing employment data of New Jersey, i.e., the treatment group, and Pennsylvania, i.e., the control group, before and after the rise. In CK and subsequent re-examinations~\citep{Lu_Rosenbaum,Card_Krueger_2020}, the number of full-time employees (FT) and the number of part-time employees (PT) in each restaurant were converted to a single number, the full-time equivalent employees (FTE), which is defined as $FTE = FT + 0.5\times PT$. The causal impact of the rise on FTE was then investigated.

This conversion may lose fine details in the characteristics of restaurants. For the same FTE, there might be a restaurant with high FT and low PT and another one with low FT and high PT, depending on each restaurant's characteristics. These restaurants might respond differently to the increase in minimum wages. Therefore, analyzing only the univariate FTE risks confounding those different trends. Simultaneous analysis of FT and PT can be expected to better capture different trends in the responses of restaurants by dissecting the causal impacts of increasing minimum wage on FT and PT. The estimation results of the $2$-dimensional counterfactual distribution of FT and PT in New Jersey after the rise are shown in Fig.~\ref{fig:ck_result}. More details on the dataset can be found in Appendix~\ref{appendix:sim_settings}. 

\begin{figure}[!h]
     \centering
    \includegraphics[width=\textwidth]{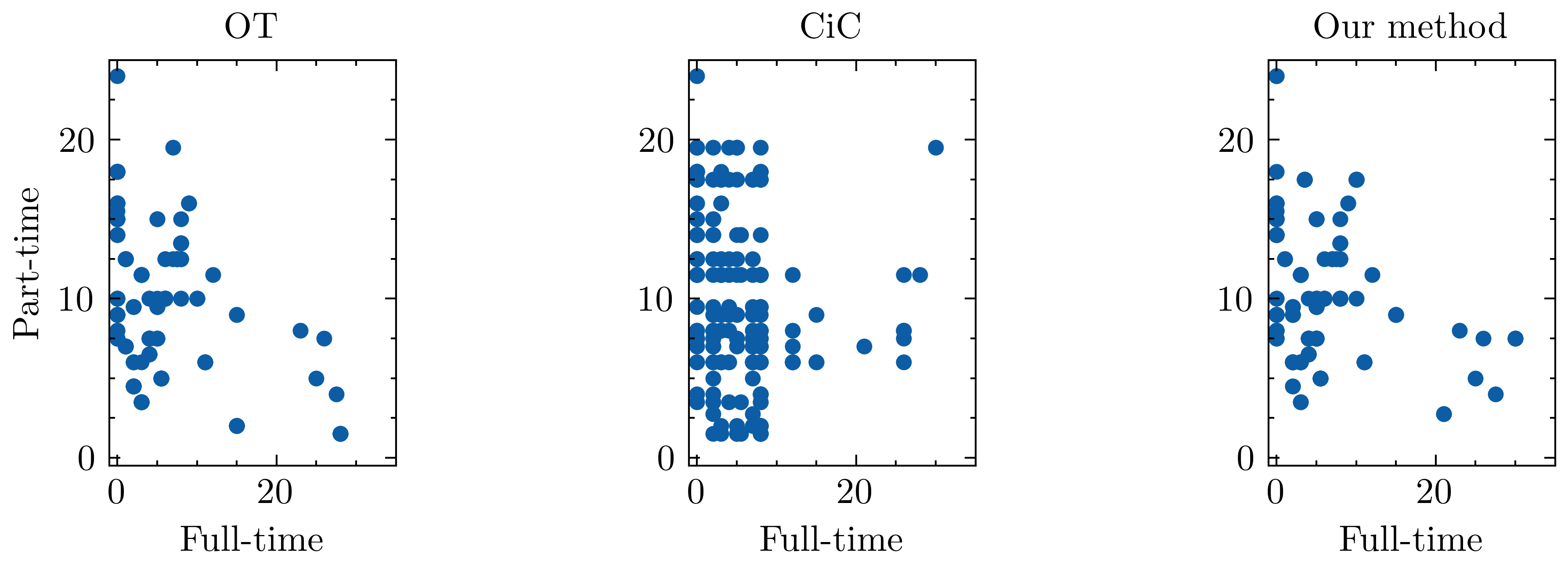}
    \caption{\textbf{Counterfactual distribution of the numbers of full-time (FT) and part-time (PT) employees in restaurants in New Jersey}. The data is from~\citet{NBERw4509}. The panel of our method shows the result of one typical run. Our method captures well the correlation between FT and PT in the result of OT, while CiC struggles in the region of high FT and PT. The OT distance between the result of CiC and the result of OT is $72.26$. For our method, this number averaged over $1000$ runs is $68.66 \pm 1.42$.} 
    \vspace{-1.5em}
        \label{fig:ck_result}
\end{figure}

We compare our method with CiC by measuring how close the estimation results of our method and CiC to the estimation result of OT, which can be regarded as the standard in terms of accuracy when one does not know the ground truth. A quick visual inspection of Fig.~\ref{fig:ck_result} reveals that our result captures well the relationship between PT and FT as found in the result of OT, while CiC struggles in the region where both PT and FT are large. Since there is randomness in our method, we measure the OT distance between our method and the result of OT, averaging over $1000$ runs. The OT distance between the result of CiC and the result of OT is $72.26$, while that of our method is $68.66 \pm 1.42$, where the confidence interval is two standard deviations. These numbers reinforce the aforementioned visual impressions and offer statistical evidence to support the conclusion that our method captured better the relationship between FT and PT than CiC.   

\section{Concluding remarks}
\label{sec: concluding}
We proposed a method for estimating the counterfactual distribution in multidimensional CiC models. Our method, like CiC, enjoys the computational efficiency of one-dimensional optimal transports while utilizing correlation information that is ignored under CiC. Through synthetic and real-dataset experiments, our method is shown to consistently outperform CiC in terms of accuracy, while running at a fraction of the time of the multidimensional OT approach. In future works, we plan to explore the robustness of our proposed method in the presence of outliers or noises, as well as in other causal settings, such as the triple difference model~\citep{Gruber_1994,Olden_2022}.

\acks{This work was partially supported by Shiga University Competitive Research Fund, JST CREST JPMJCR22D2, JPMJCR2015, JST MIRAI program JPMJMI21G2, and ISM joint-research 2023-ISMCRP-2010.}

\bibliography{ML_CI_library.bib, bibEPT21, bibSobolev22}

\appendix
\renewcommand\thefigure{\thesection.\arabic{figure}}    
\setcounter{figure}{0}   

\section{Details on experiment settings}
\label{appendix:sim_settings}
\subsection{The illustrative examples}
The latent distributions in two examples are as follows. In the bivariate Gamma example, the first dimension of $\nu^{\texttt{C}}$ is a Gamma distribution with shape $2$ and scale $3$, while the second dimension is Gamma with shape $3$ and scale $2$. For $\nu^{\texttt{T}}$, the first and second dimensions are reversed. In the Gaussian mixture example, the first and second dimensions of $\nu^{\texttt{C}}$ are  $0.5\mathcal{N}(1,1) + 0.5\mathcal{N}(5,1)$ and $0.5\mathcal{N}(2,1) + 0.5\mathcal{N}(4,1)$, respectively. For $\nu^{\texttt{T}}$, the first and second dimensions are reversed.

The production functions $h_0$ and $h_1$ are  
$h_0(u)  = \big(\begin{smallmatrix}
  1 & 0.5 \\
  0.5 & 1
\end{smallmatrix}\big)u$ and $h_1(u) = \big(\begin{smallmatrix}
  1 & -0.5 \\
  -0.5 & 1
\end{smallmatrix}\big)u$ for a length-$2$ vector $u$. These functions are co-monotone, and thus the natural drift map $\texttt{f}$ is identifiable~\citep{torous_2021}.  

\subsection{Settings for the experiments with varying $d$}

We discuss the settings for the production functions $h_0$ and $h_1$. When $d \ge 2$, for the natural drift map $\texttt{f}$ to be identifiable, the functions $h_0$ and $h_1$ need to be co-monotone~\citep{torous_2021}, i.e.,
\[
\langle h_0(x) - h_0(y), h_1(x) - h_1(y)  \rangle \ge 0, \quad \forall x,y\in \RR^d.
\]
When $h_0(x) = \mathbf{H}_0x$ and $h_1(x) = \mathbf{H}_1x$ for $\RR^{d \times d}$ matrices $\mathbf{H}_0$ and $\mathbf{H}_1$, this condition is satisfied if $\mathbf{H}_0^T\mathbf{H}_1$ is positive semi-definite. We generate one matrix $\mathbf{H}_0$ as a $d \times d$ matrix where each off-diagonal entry is uniformly distributed in $(0,1)$ and the diagonal entries are $1$. Note that $\mathbf{H}_0$ generated this way is almost surely invertible. We then generate a diagonal matrix $\mathbf{B}$ where each diagonal entry is uniformly distributed in $(0,1)$. We then let $\mathbf{H}_1 = (\mathbf{H}_0^{-1})^T\mathbf{B}$. This will ensure that $\mathbf{H}_0^T\mathbf{H}_1$ is equal to $\mathbf{B}$ and thus positive semi-definite. We then fix the pair ($\mathbf{H}_0$, $\mathbf{H}_1$) and then generate datasets using this pair.

The latent distributions are as follows. For the control group, each dimension independently follows a Gamma distribution with shape $2$ and scale $3$. For the treatment group, each dimension independently follows a Gamma distribution with shape $3$ and scale $2$.

\subsection{CK data}
The dataset is downloaded from \url{https://davidcard.berkeley.edu/data_sets/njmin.zip}. We also include the following covariates into the analysis: HRSOPEN, OPEN, NMGRS, NREGS, INCTIME, PSODA, and PENTREE, and estimate the $9$-dimensional counterfactual distribution. We process the data by removing samples that contain missing values at any covariates. The final numbers of samples after this pre-processing are $57$ for the control group and $220$ for the treatment group.  

\section{Additional results with different hyperparameters of Sinkhorn}
\label{appendix:sinkhorn}
In Sinkhorn, an entropic regularization term $\lambda \Omega(T)$, where $\Omega(T)$ is the entropy of the transportation plan $T$, is added to the objective function of OT. The hyperparameter $\lambda \ge 0$ controls the strength of the regularization. The results reported in Figs.~\ref{fig:experiment_varying_n} and~\ref{fig:experiment_varying_d} are obtained using $\lambda = 30$. We report in the following figures two more cases when $\lambda = 10$ and $\lambda = 90$. 
\begin{figure}[!h]\centering
     \centering
\includegraphics[width = \textwidth]{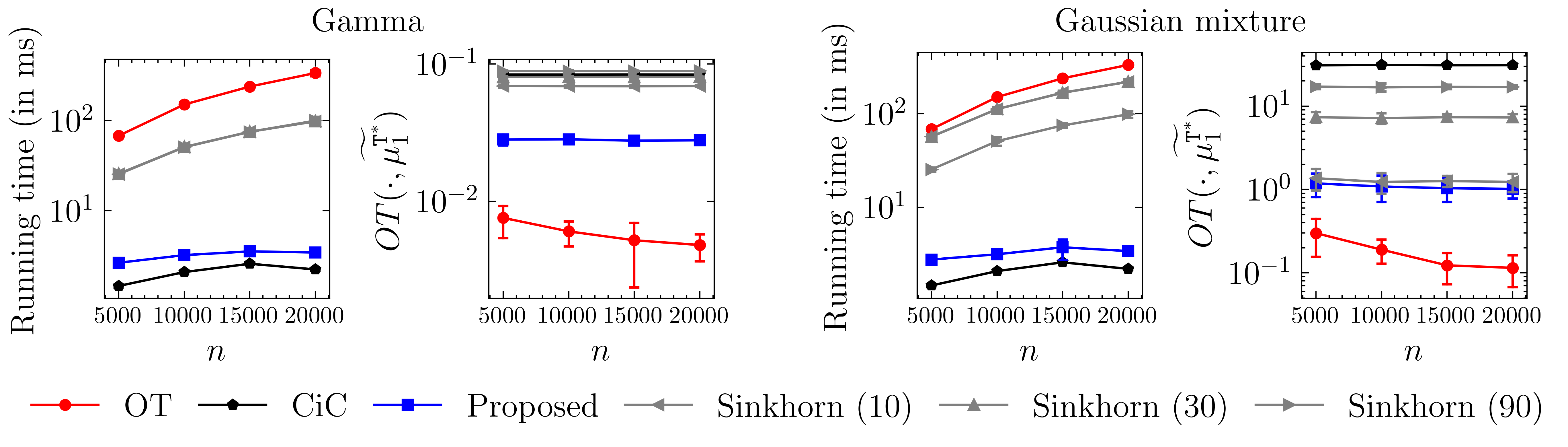}
         \caption{Additional results of Sinkhorn for the case of varying the number of samples $n$ while fixing the dimension $d = 2$.}
        \label{fig:experiment_varying_n_appendix}
\end{figure}
\begin{figure}[!h]
     \centering
         \includegraphics[width=0.7\textwidth]{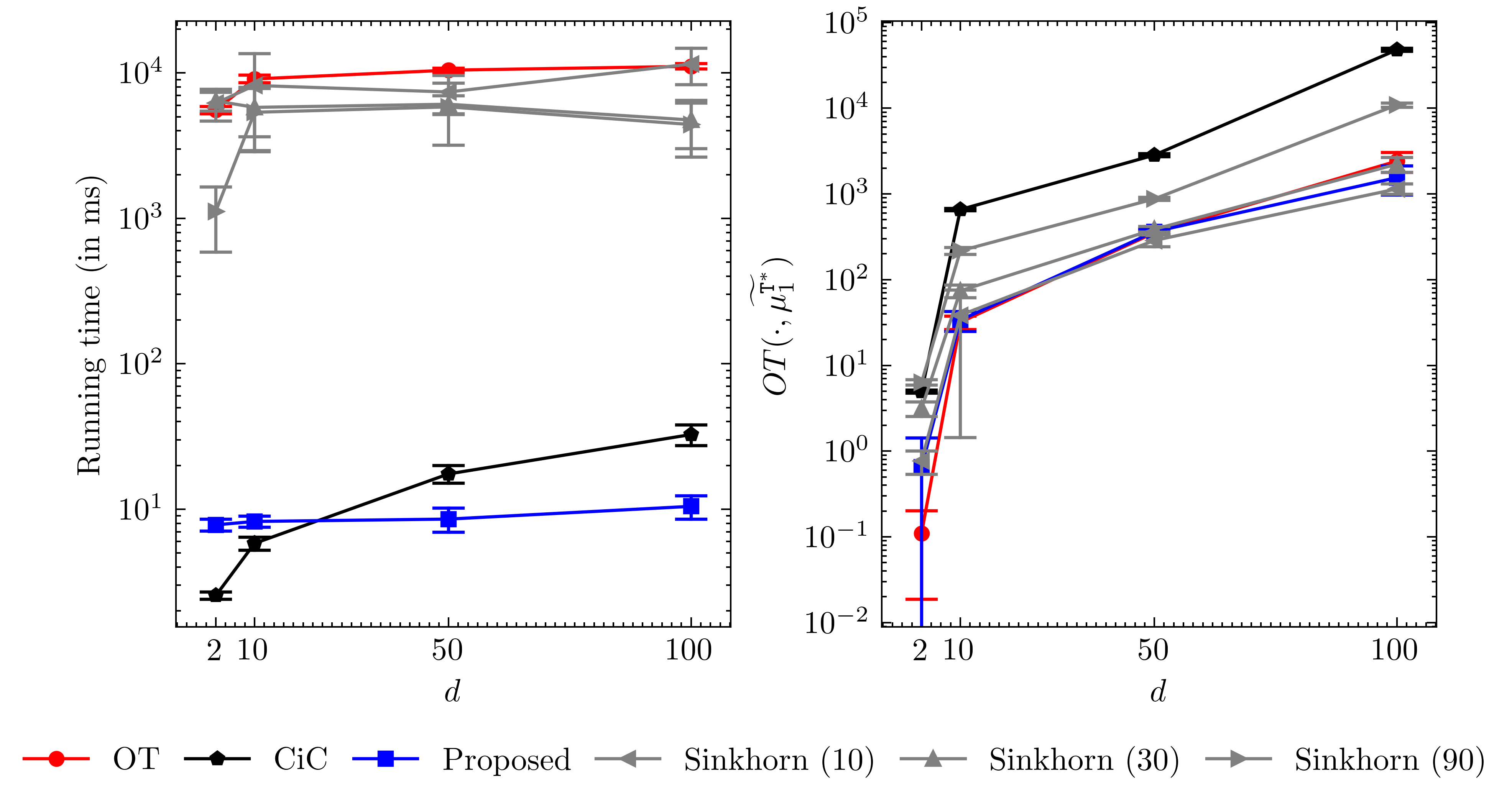}
         \caption{Additional results of Sinkhorn for the case of varying the dimension $d$ while fixing number of samples $n = 5000$.}
        \label{fig:experiment_varying_d_appendix}
\end{figure}

\section{Brief review}\label{sec:supp_review_OT}
In this section, we further give brief review for some technical details on optimal transport (OT) which are used in our work.

\paragraph{Kantorovich formulation of OT.} Given two probability distributions $\mu, \nu$ with a cost function $c$, the Kantorovich formulation of OT is as follow:
\[
\text{OT}(\mu, \nu) = \inf_{\pi \in \Pi(\mu, \nu)} \int_{\RR^d \times \RR^d} \pi(x, y) c(x, y) d\mu(x) d\nu(y),
\]
where $\pi$ is known as the transport plan, and $\Pi(\mu, \nu)$ is the set of all probability distributions on the product space $\RR^d \times \RR^d$ such that its first and second marginals equal to $\mu, \nu$ respectively.

\paragraph{Optimal condition for OT with $d=1$.} If $\mu, \nu$ are two probability measures supported on $\RR$, and $\mu$ is atomless (i.e., $\mu$ is absolutely continuous with respect to the Lebesgue measure), then there exists at least a transport map $T$ such that $T_{\sharp}\mu = \nu$~\citep[Lemma 1.27]{SantambrogioBook}. With quadratic cost, the transport map will be the derivative of a convex function, i.e., a nondecreasing map (see~\citep[Remark 1.23]{SantambrogioBook} and~\citep[\S2]{gangbo1999monge} for further details).

\paragraph{Optimal condition for OT with $d>1$.} It is also known as Brenier theorem~\citep{brenier1991polar}.

Given two probability measures $\mu, \nu$ supported on the $\RR^d$ space such that $\mu$ is absolutely continuous with respect to the Lebesgue measure. Then, for all possible map $\widetilde{T}: \RR^d \mapsto \RR^d$ such that $\widetilde{T}_{\sharp}\mu = \nu$, there is an unique Brenier map $T$ which is the gradient of a convex function, and $T$ is the optimal transport in the following sense: the Kantorich formulation of OT between $\mu$ and $\nu$ admits a unique optimal transportation plan $\pi^*$ such that $(x, y) \sim \pi^*$ if and only if $x \sim \mu$ and $y = T(x)$, $\mu$-a.s. For further details, please see~\citep[Theorem 1.17]{SantambrogioBook} and~\citep{brenier1991polar, mccann1995existence}.

\section{Further discussions}

\paragraph{About the max-min robust OT.} Max-sliced Wasserstein~\citep{Deshpande_2019_CVPR} is similar to our proposed method, i.e., they consider the sphere for the uncertainty set $\Omega$ of projections. However, when $\Omega$ is a sphere, the problem becomes non-convex and non-smooth. The entropic regularized OT~\citep{Cuturi-2013-Sinkhorn} is a popular approach to reduce the computation of OT into quadratic computation complexity, but it comes with a trade-off for a dense optimal transport plan.

\paragraph{About the causal model.} The causal model described in this paper is the same as that of~\citet{athey_imbens_2006} when $d = 1$ and that of~\citet{torous_2021} for a general $d$. When $d = 1$, it is a non-linear extension of the classical DiD model~\citep{ANGRIST19991277,BLUNDELL19991559}. Another notable non-linear extension of this DiD model is provided in~\citet{Abadie_2006}. Some stronger assumptions to model the natural drift have been proposed in, for example,~\citet{Callaway_2019},~\citet{Roth_2023}, and~\citet{Bonhomme_2011}.

\paragraph{About the use of machine learning methods in causal inference and causal discovery}
Recently, \cite{tu2022optimal} provided the first method that uses optimal transport for solving causal discovery, i.e., the task of uncovering the graph that represents the dependence relationships between variables. \cite{akbari2023learning} generalized the method to higher dimensions and different noise settings. Other machine learning tools and frameworks have also been used in causal inference/causal discovery, such as neural networks~\citep{hwang2023on,ashman2023causal,Immer_2023,kladny2023deep}, the Kullback-Leibler divergence~\citep{pmlr-v213-wildberger23a}, and reinforcement learning and meta learning~\citep{sauter2023a}.

\end{document}